\documentclass[10pt,twocolumn,letterpaper]{article}

\usepackage{cvpr}

\usepackage{amsmath,amsfonts,bm}

\def\eqref#1{equation~\ref{#1}}

\def\1{\bm{1}}

\DeclareMathAlphabet{\mathsfit}{\encodingdefault}{\sfdefault}{m}{sl}
\SetMathAlphabet{\mathsfit}{bold}{\encodingdefault}{\sfdefault}{bx}{n}

\usepackage{url}
\usepackage{amsmath}
\usepackage{multicol}
\usepackage{multirow}
\usepackage{booktabs}
\usepackage{bbding}
\usepackage{graphicx}
\usepackage{float}
\usepackage{stfloats}
\usepackage{pifont}
\usepackage{makecell}
\usepackage{microtype}
\newcommand{\dui}{\ding{51}}
\newcommand\chn[1]{}

\definecolor{cvprblue}{rgb}{0.21,0.49,0.74}
\usepackage[pagebackref,breaklinks,colorlinks,allcolors=cvprblue]{hyperref}

\def\paperID{12468} 
\def\confName{CVPR}
\def\confYear{2025}

\title{AniMer: Animal Pose and Shape Estimation Using Family Aware Transformer}

\author{ Jin Lyu$^{1,*}$, Tianyi Zhu$^{2,*}$, Yi Gu$^{3}$, Li Lin$^{1,4}$, Pujin Cheng$^{1,4}$, Yebin Liu$^{5}$, Xiaoying Tang$^{1,\dag}$, Liang An$^{5,\dag}$\\
$^{1}$Southern University of Science and Technology \\
$^{2}$China Mobile Research Institute \\
$^{3}$The Hong Kong University of Science and Technology \\
$^{4}$The University of Hong Kong $^{5}$Tsinghua University \\
}

\begin{document}

\maketitle
\let\thefootnote\relax\footnotetext{* Equal contribution.}
\let\thefootnote\relax\footnotetext{$^\dag$ Corresponding author.}

\begin{abstract}

Quantitative analysis of animal behavior and biomechanics requires accurate animal pose and shape estimation across species, and is important for animal welfare and biological research. However, the small network capacity of previous methods and limited multi-species dataset leave this problem underexplored. To this end, 
this paper presents \textbf{AniMer} to estimate \textit{\textbf{ani}}mal pose and shape using family aware Transfor\textit{\textbf{mer}}, enhancing the reconstruction accuracy of diverse quadrupedal families. 
A key insight of AniMer is its integration of a high-capacity Transformer-based backbone and an animal family supervised contrastive learning scheme, unifying the discriminative understanding of various quadrupedal shapes within a single framework. 
For effective training, we aggregate most available open-sourced quadrupedal datasets, either with 3D or 2D labels. To improve the diversity of 3D labeled data, we introduce CtrlAni3D, a novel large-scale synthetic dataset created through a new diffusion-based conditional image generation pipeline. CtrlAni3D consists of about 10k images with pixel-aligned SMAL labels. In total, we obtain 41.3k annotated images for training and validation. 
Consequently, the combination of a family aware Transformer network and an expansive dataset enables AniMer to outperform existing methods not only on 3D datasets like Animal3D and CtrlAni3D, but also on out-of-distribution Animal Kingdom dataset. Ablation studies further demonstrate the effectiveness of our network design and CtrlAni3D in enhancing the performance of AniMer for in-the-wild applications. 
Project page:~\url{https://luoxue-star.github.io/AniMer_project_page/}.

\end{abstract}
\section{Introduction}

Animal pose and shape estimation from images is essential for capturing animal behavior, biomechanics, and interactions with their environment, thereby yielding vital insights for animal welfare, agriculture, ecology and life sciences. The integration of geometric and appearance information from diverse species into a unified deep neural network represents a significant step to comprehensively interpret the intricate and dynamic animal world. 
A notable endeavor in this domain involves the estimation of pose and shape parameters from an articulated animal template known as the SMAL model~\citep{zuffi20173d}, primarily targeting general quadrupeds. While extensive researches have been conducted regarding single species or families, such as horses~\citep{zuffi2024varen,li2021hsmal,zuffi2019three} and dogs~\citep{Sabathier2024AnimalAR,ruegg2023bite,rueegg2022barc,li2021coarse,biggs2020wldo,biggs2019creatures}, investigations into other quadrupedal species, including cats, cows or hippos, remain relatively underexplored.

The reconstruction of multiple species within a single network is hampered by the limited capacity of backbones and the scarcity of multi-species datasets annotated with SMAL labels. Recent advancements in human mesh recovery via the SMPL model~\citep{SMPL:2015} have proved that using a high-capacity backbone in conjunction with large-scale datasets significantly enhances the accuracy of human pose and shape estimation in diverse settings~\citep{goel2023humans,pavlakos2024reconstructing,cai2024smpler}. However, this simple and effective paradigm remains untested within the context of animal studies. \citet{xu2023animal3d} introduce the first large-scale dataset Animal3D featuring SMAL mesh annotations; nevertheless, their study predominantly utilizes CNN-based networks such as HMR~\citep{kanazawa2018end} and WLDO~\citep{biggs2020wldo}.

\begin{figure*}[ht]
    \centering
    \includegraphics[width=0.85\linewidth]{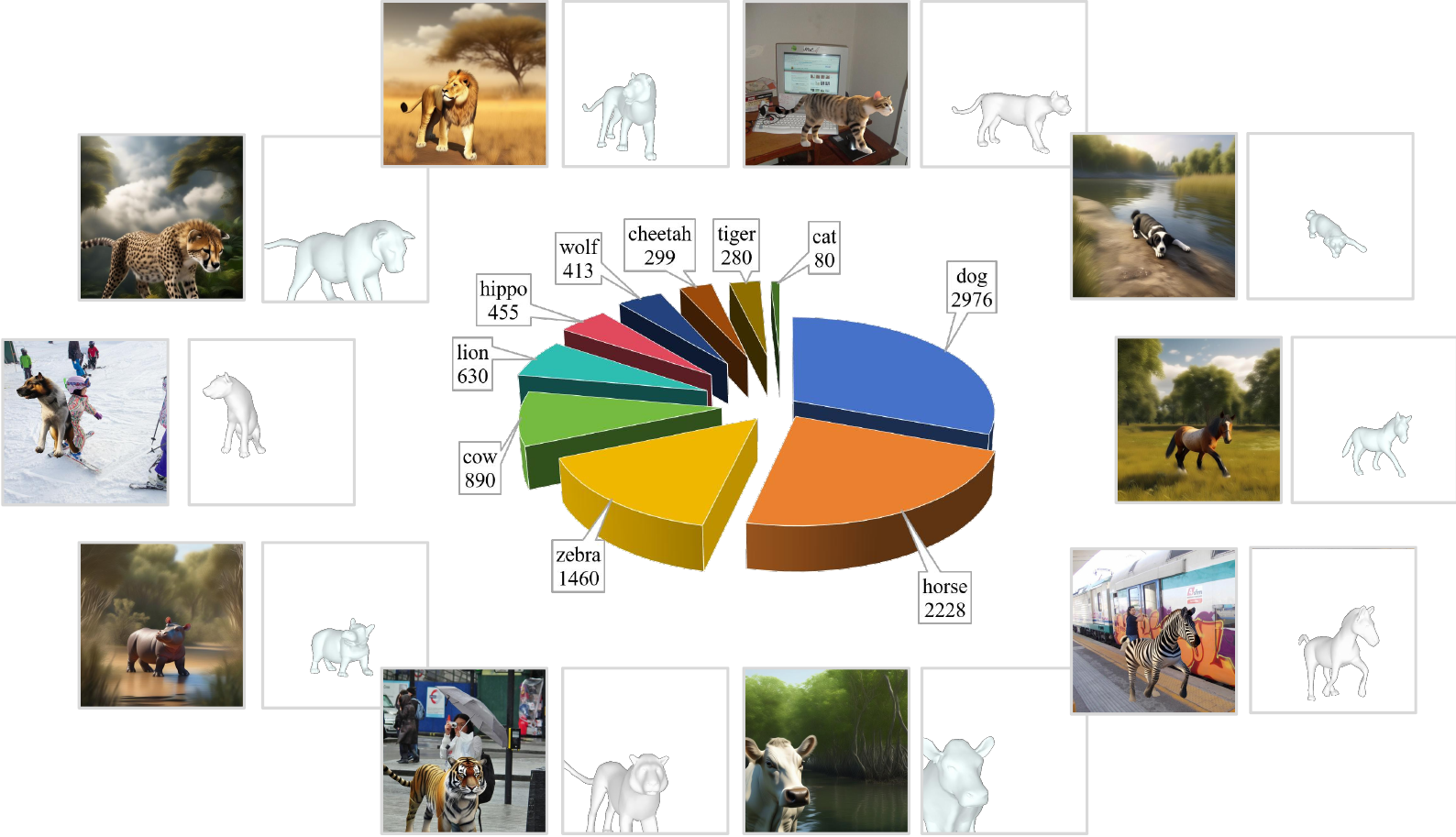}
    \caption{\textbf{CtrlAni3D dataset statistics and visual samples.} Ten samples are shown for ten species in the same order of pie plot. For each image pair, the left side displays the generated animal image whose background comes from either COCO ~\citep{lin2014microsoft} or AI-synthesis, while the right side presents the rendering of the SMAL mesh label. Note that the synthesis process naturally considers generating truncated images. }
    \label{fig:dataset_samples}
\end{figure*}

In this paper, we present \textbf{AniMer}, a systematic approach pursuing accurate \textit{\textbf{ani}}mal pose and shape estimation using Transfor\textit{\textbf{mer}} which surpasses existing methodologies on all benchmarks.
This success relies on two key scaling factors: scaled backbone and scaled dataset. In previous animal researches, the well-known scaled Transformer backbone proposed by ViT~\citep{xu2023vitpose++} has only been used for 2D animal pose estimation. 
We go beyond it by connecting ViT with a Transformer-based decoder to yield SMAL parameters, following HMR2.0~\citep{goel2023humans}. Different from HMR2.0, our Transformer is family aware by applying a new animal family supervised contrastive learning scheme, which enhances the discrimination of animal shapes through class token. Moreover, due to the limited available data space and prior knowledge of animals compared to that of human, we change the residual parameter decoding used by HMR~\citep{kanazawa2018end} and HMR2.0~\citep{goel2023humans} to direct parameter decoding, and adopt a two-stage training scheme to train on 3D dataset first instead of training on full dataset directly. 

Acquiring a large number of animal images with full 3D annotations is difficult, and synthetic datasets can mitigate the scarcity of real data. The plausibility of traditional CG-generated animal images is hindered by the coarse texture quality and sophisticated lighting and shadow control. 
These limitations motivate our usage of ControlNet~\citep{zhang2023adding} to create high quality AI hallucinated images conditioned on SMAL structures. 
This principle effectively bridges the domain gap between parametric labels and images, and help us to create a novel large scale dataset named CtrlAni3D with minimal human labor involved. 
Specifically, we prompt ControlNet with textual descriptions of animal behaviors and rendered SMAL mask and depth images, resulting in highly realistic visual outputs, as illustrated in Fig.~\ref{fig:dataset_samples}. To ensure dataset quality, we use SAM2.0~\citep{ravi2024sam2} together with manual verification to filter implausible generated images. Finally, CtrlAni3D comprises 9711 images annotated with pixel-aligned SMAL mesh, and its scalable nature allows for further expansion. Finally, we obtain scaled dataset for AniMer training by aggregating most available open-sourced quadrupedal datasets and our CtrlAni3D, resulting in a comprehensive set of 41.3k images annotated with either 3D mesh or 2D keypoints. 


To rigorously evaluate the efficacy of the AniMer model and the CtrlAni3D dataset, we conduct extensive empirical studies. Our findings indicate that AniMer, when trained on the same full multi-species datasets, significantly outperforms CNN-based methods such as HMR and WLDO~\citep{biggs2020wldo}, which serve as baselines in the Animal3D research. Besides, compared with HMR2.0, our animal family supervised contrastive learning scheme improves the pose and shape estimation precision on all benchmarks. 
Through comprehensive ablation studies, we verify that CtrlAni3D enhances the generalization abilities of AniMer on the out-of-distribution (OOD) Animal Kingdom dataset~\citep{Ng_2022_CVPR}, which is unseen during training.


\section{Related Works}

\textbf{Animal Pose and Shape Estimation.}
Compared to keypoint-based animal pose estimation methods~\cite{mathis2018deeplabcut,pereira2022sleap,ye2024superanimal,dabhi20243d}, surface-based methods could provide both shape and pose information simultaneously.
In this paper, we focus on template-based methods instead of template-free methods~\citep{yang2022banmo,yao2022lassie,jakab2024farm3d,li2024learning}. 
Reconstructing shape templates has been studied in various types of animal families such as birds~\citep{wang2021birds}, mice~\citep{an2023three}, non-human primates~\citep{neverova2020continuous} and quadrupeds~\citep{zuffi2018lions}. For quadrupeds, \citet{zuffi20173d} proposes the well-known SMAL, which is built upon 41 scans of animal toys. Due to the limited geometry accuracy of SMAL for representing specific species, most previous methods stride over predicting SMAL parameters only and further enhance the geometry of horses and dogs. For example, ~\citet{li2021hsmal} designs the horse-specific hSMAL model and applies it to the problem of video lameness detection. VAREN~\citep{zuffi2024varen} further improves hSMAL with high quality horse scans. Similarly, the use of SMAL for dog mesh recovery is addressed by adding bone lengths control~\citep{biggs2020wldo}, by adding per-vertex deformation~\citep{li2021coarse}, by introducing breed loss~\citep{rueegg2022barc}, by ground contact constraints~\citep{ruegg2023bite}, or by temporal avatar optimization~\citep{Sabathier2024AnimalAR}. Though high quality reconstruction has been achieved on horses and dogs, end-to-end SMAL estimation has not been fully addressed on other challenging species. Animal3D~\citep{xu2023animal3d} provides the first large scale 3D benchmark for general quadruped SMAL estimation, yet neglects in-depth research on the network design and training. 


\noindent\textbf{Transformer Based Human Mesh Recovery.}
The Transformer architecture~\citep{vaswani2017attention} has revolutionized the field of Natural Language Processing (NLP) by enabling unprecedented accuracy and efficiency in a wide range of tasks. 
Inspired by its success in NLP, one of the core part of Transformer, i.e. self-attention, has been widely used for human mesh recovery~\citep{kocabas2021pare, wan2021encoder, shen2023global, shin2024wham}. ~\citet{dosovitskiy2020image} proposes Vision Transformer(ViT), which divides an image into patches as the input to Transformer. ViT has achieved state-of-the-art performance on several computer vision tasks including human mesh recovery using SMPL~\citep{kocabas2021pare, wan2021encoder, shen2023global, goel2023humans, shin2024wham}. Among all these works, HMR2.0~\citep{goel2023humans} is a milestone which demonstrates the effectiveness of simply using ViT backbone and large scale datasets to achieve accurate mesh recovery and in-the-wild generalization ability. Inspired by this, HaMeR~\citep{pavlakos2024reconstructing} employs ViT backbone to achieve highly accurate hand mesh recovery, which is further extended to interacting hands~\citep{lin20244dhands}. Similarly, SMPLer-X~\citep{cai2024smpler} scales up expressive human pose and shape estimation using ViT backbone and the combination of 32 datasets. Although impressive results have been achieved in human mesh recovery, the effect of ViT backbone for animal pose and shape estimation remains unexplored. 


\noindent\textbf{Synthetic Animal Training.}
Compared to human pose estimation which benefits from large-scale datasets, acquiring annotated images of animals is significantly more difficult. Therefore, synthetic dataset would alleviate this problem by rendering the input and output simultaneously. Most previous methods only attempt to render RGB images~\citep{cao2019cross,mu2020learning,plum2023replicant,ruegg2023bite,shooter2024digidogs,xu2023animal3d} or depth images~\citep{Kearney_2020_CVPR} using traditional computer graphics pipelines, ignoring the image hallucination ability of generative AI models such as stable diffusion~\citep{rombach2022high} or ControlNet~\citep{zhang2023adding}. 
Although traditional CG rendering achieves success in assisting network training, large scale high quality CG assets are expensive to obtain, and the plausibility of rendered images is hindered the sophisticated lighting and shadow control. In contrast, ~\citet{ma2023generating} uses 3D
visual prompts and large language model (LLM) text prompts to generate diverse images from a static CAD model, pioneering a new direction. We extend their idea to animal image synthesis from dynamic SMAL animations to construct the CtrlAni3D dataset. The most similar concurrent work to us is SPAC-Net~\citep{jiang2023spac} which performs style-transfer from synthetic domain to real domain using ControlNet~\citep{zhang2023adding}. However, they rely on textured CAD assets instead of untextured SMAL, making it difficult to assist SMAL pose and shape estimation. 
\begin{figure*}[ht]
    \centering
    \includegraphics[width=\linewidth]{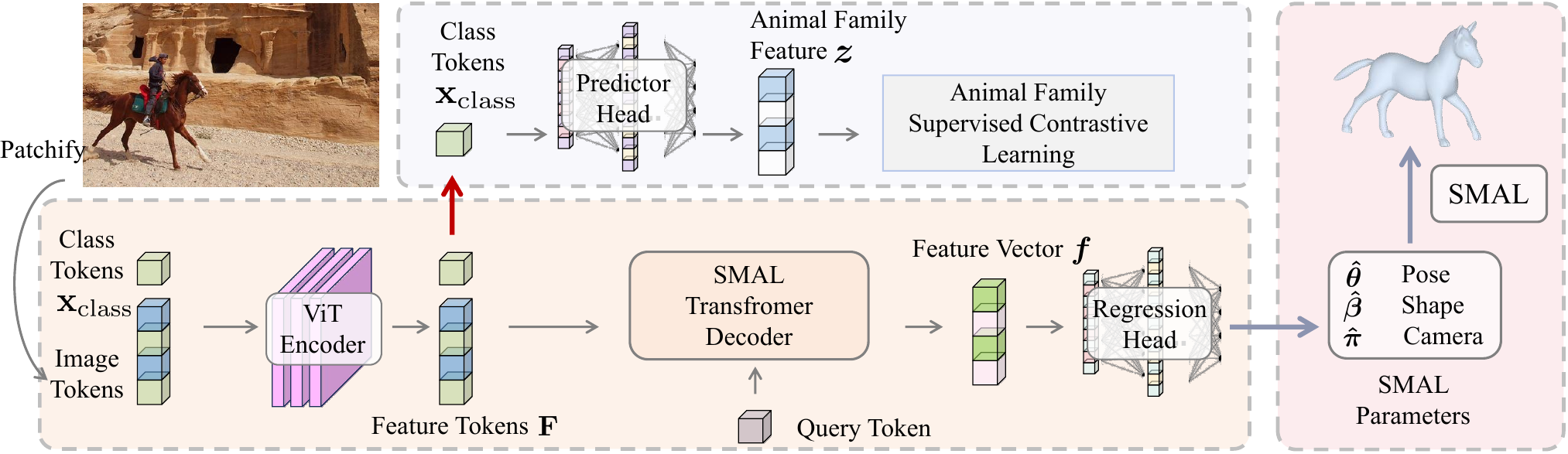}
    \caption{\textbf{AniMer network architecture.} AniMer consists of (1) a ViT encoder that extracts image features; (2) a transformer decoder that processes the image features from the encoder; (3) a predictor head (MLPs) that generates animal family feature for supervised contrastive learning; and (4) a regression head (MLPs) that estimates the shape \(\hat{\beta}\), pose \(\hat{\theta}\), and camera parameters \(\hat{\pi}\).}
    \label{fig:pipeline}
\end{figure*}

\section{Reconstructing Animals Using AniMer}
\subsection{Preliminaries}
\noindent\textbf{The SMAL Model.} The SMAL model, denoted as \(\mathcal{M}(\beta, \theta, \gamma)\), is a 3D parametric shape model designed for quadrupeds. The shape parameter \(\beta\ \in \mathbb{R}^{41} \) is derived from 41 3D scans of various animal figurines, including cats, dogs, horses, cows, and hippos~\citep{zuffi20173d}. The pose parameter \(\theta \in \mathbb{R}^{35 \times 3}\) represents the rotation of each joint relative to its parent joint, expressed in terms of axis-angle. Controlled by \(\beta\) and \(\theta\), the SMAL model outputs a mesh with vertices \(V \in \mathbb{R}^{3889 \times 3}\) and faces \(F \in \mathbb{N}^{7774 \times 3}\) through linear blend skinning (LBS) process. The animal body joints are regressed from vertices by \(J \in \mathbb{R}^{35 \times 3} = W \cdot V\), where $W\in\mathbb{R}^{35\times3889}$ represents a linear mapping matrix.

\noindent\textbf{Camera Projection.} Following ~\citet{goel2023humans}, \(\pi(\cdot)\) represents the projection process of a weak-perspective camera model, determined by a translation vector $T\in\mathbb{R}^3$, a fixed focal length $f=1000$ and thereby a fixed intrinsic matrix $K$. The global orientation $R$ is the rotation of SMAL root joint, therefore we ignore it here. Consequently, a 3D point $X$ is projected as 2D point $x$ by $x=\pi(X)=\Pi(K(X+T))$, where $\Pi$ converts homogeneous coordinates $(u,v,d)^T$ to pixel coordinates $(u/d,v/d)^T$.

\subsection{The Architecture of AniMer}
The full architecture of AniMer is shown in Fig.~\ref{fig:pipeline}. 
Given an image \(I \in \mathbb{R}^{H \times W \times 3}\) and a class token \(\mathbf{x}_{class} \in \mathbb{R}^{1 \times 1280}\), we first utilize a ViT encoder to extract image feature tokens \(\mathbf{F} \in \mathbb{R}^{192 \times 1280}\), while the class token interacts with the image to capture information about the animal family. We then feed the feature tokens \(\mathbf{F}\) into a SMAL transformer decoder to obtain a feature vector \(\boldsymbol{f} \in \mathbb{R}^{1 \times 1024}\). Finally, independent multi-layer perceptrons (MLPs) are used to predict the shape parameters \(\hat{\beta}\), pose parameters \(\hat{\theta}\), and camera parameters \(\hat{\pi}\), where $\hat{\cdot}$ means estimated parameters. At the same time, the class token is fed into the predictor head for animal family supervised contrastive learning, which will be detailed in Sec.~\ref{sec:sec:const}. Note that the weights of ViT encoder are pre-trained using ~\citet{xu2023vitpose++}, which significantly enhances mesh recovery performance.

Despite the class token, 
AniMer features two key differences compared with previous HMR2.0~\citep{goel2023humans} and HaMeR~\citep{pavlakos2024reconstructing}. 
First, both HMR2.0 and HaMeR regress the residual SMPL/MANO parameters with respect to the non-zero mean parameters computed from large scale motion databases. In contrast, we choose to directly decode the final SMAL parameters due to limited SMAL pose prior. Second, both HMR2.0 and HaMeR train on the whole datasets in single stage. Instead, we train AniMer in two stages. 
At the first stage, we train AniMer using only 3D data to ensure the network could predict feasible shapes and poses. At the second stage, all 3D and 2D data are introduced for training. 
The insight is that the size of 3D datasets for animal is much smaller than that of human at present, resulting in an imbalanced 3D and 2D data scale.  
Unless otherwise specified, we train two stages for 500 and 700 epochs respectively.  

\subsection{Animal Family Supervised Contrastive Learning}
\label{sec:sec:const}
Different from human whose SMPL shape parameters come from the same multivariate normal distribution, animals demonstrate at least two levels of shape differences: inter-family level and intra-family level. For example, dogs share similar shapes between each other, yet share distinct shapes with cows. 
To capture such two levels of shape distributions in AniMer, we propose an animal family loss based on supervised contrastive learning~\citep{khosla2020supervised}. 

Specifically, we employ a learnable class token~\cite{dosovitskiy2020image} to represent the animal family. First, this token \(\mathbf{x}_{class}\) together with a minibatch of images \(I \in \mathbb{R}^{B \times 3 \times H \times W}\) are fed into the ViT encoder. Typically, $B=16$. Then, an MLP head is utilized to generate the animal family feature \(\boldsymbol{z}\) from the class token. Finally, we directly apply supervised contrastive learning to \(\boldsymbol{z}\) through loss $\mathcal{L}_{\text{con}}$ defined as follows:
\begin{equation}
    \label{animal family loss}
    \mathcal{L}_{\text{con}} =   \sum_{i \in I} 
                            \frac{-1}{|P(i)|} 
                            \sum_{p \in P(i)} \frac{\exp{(\boldsymbol{z_i} \cdot \boldsymbol{z_p} / \tau) }}
                            {\sum_{o \in O(i)} \exp{(\boldsymbol{z_i} \cdot \boldsymbol{z_o} / \tau)}}. 
\end{equation}
Within a minibatch, \(P(i)\) represents the samples that share the same family label of $i$, $O(i)$ represents the samples other than $i$. The notation \(|P(i)|\) denotes the size of this set. \(\tau \in \mathbb{R}^{+}\) is a scalar temperature parameter. 

\subsection{Loss Functions}
To align animal images with reconstruction results, we train our model using a comprehensive loss function that incorporates various 2D and 3D annotations. We define the main loss function \(\mathcal{L}_{\text{total}}\) as a weighted sum of several loss components, each focusing on different aspects of the model's performance. The main loss function is given by
\begin{equation}
\begin{aligned}
    \label{total_loss}
    \mathcal{L}_{\text{total}} =& \lambda_{\text{3D}} \mathcal{L}_{\text{3D}} + \lambda_{\text{2D}} \mathcal{L}_{\text{2D}} + \lambda_{\text{prior}} \mathcal{L}_{\text{prior}} + \lambda_{\text{adv}} \mathcal{L}_{\text{adv}} \\ 
    + &\lambda_{\text{con}} \mathcal{L}_{\text{con}}. 
\end{aligned}
\end{equation}
Here, \(\lambda_{\text{3D}}=0.05, \lambda_{\text{2D}}=0.01, \lambda_{\text{prior}}=0.001, \lambda_{\text{adv}}=0.0005, \lambda_{\text{con}}=0.0005\) are the loss weights. For 3D training data, all losses are used. For samples without 3D annotations, the 3D loss \(\mathcal{L}_{\text{3D}}\) is disabled.

\textbf{3D Loss.} For images annotated with SMAL model parameters \(\beta\) and \(\theta\), we supervise these parameters directly to enable fast convergence. We also supervise the estimated 3D keypoints \(\hat{K}_{3D}\) with ground truth \(K_{3D}\) to achieve better 3D joint localization. The full 3D loss function is
\begin{equation}
    \label{3D_loss}
    \mathcal{L}_{\text{3D}} = \lambda_{\beta} ||\hat{\beta} - \beta||_{2}^{2} + \lambda_{\theta} ||\hat{\theta} - \theta||_{2}^{2} + ||\hat{K}_{3D} - K_{3D}||_{1}, 
\end{equation}
where \(\lambda_{\beta}=0.01\) and \(\lambda_{\theta}=0.2\) are loss weights. \( ||\cdot||_{2}^{2} \) denotes squared L2 norm, while \( ||\cdot||_{1} \) represents L1 norm. 
\chn{T去哪了}

\textbf{2D Loss.} Most training data only contain 2D-level annotations such as 2D keypoints or masks. For these data, we supervise 2D keypoints during training using 
\begin{equation}
    \label{2D_loss}
    \mathcal{L}_{\text{2D}} = ||\pi(\hat{K}_{3D}) - K_{2D}||_{1}. 
\end{equation}
Note that, masks are only used for evaluation.

\begin{figure*}[ht]
    \centering
    \includegraphics[width=0.95\linewidth]
    {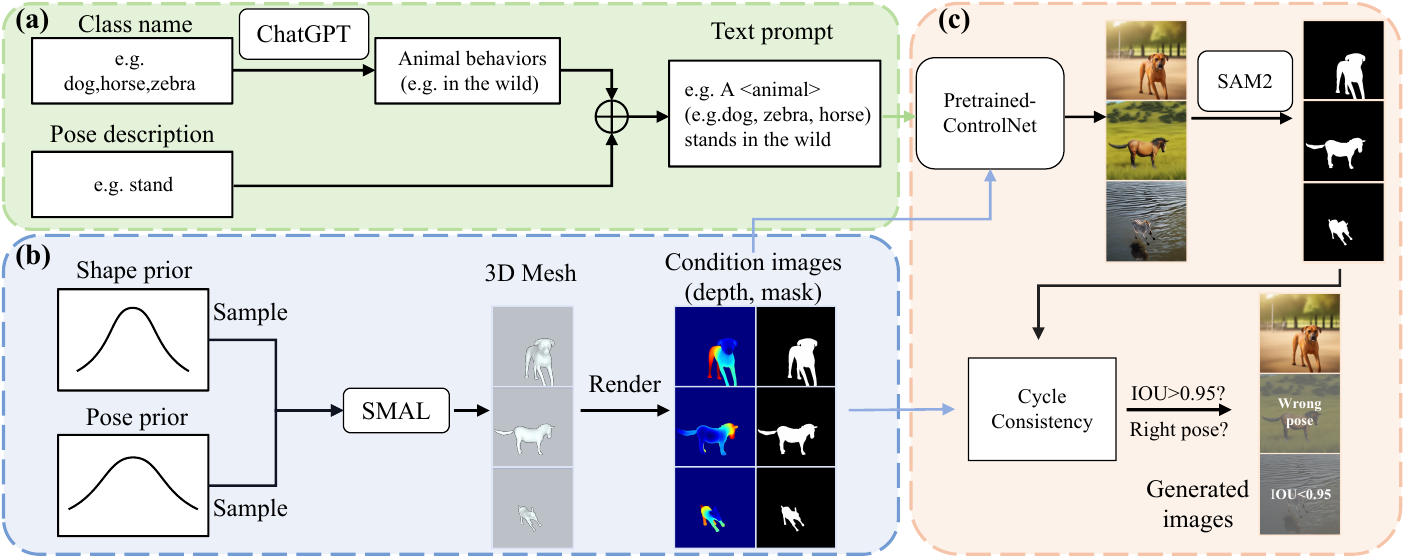}
    \caption{\textbf{CtrlAni3D dataset generation pipeline.} The whole pipeline contains three parts: \textbf{(a)} Text prompt generation. 
    \textbf{(b)} Conditional image generation. 
    \textbf{(c)} Image generation and post-processing. 
    }
    \label{fig:ctrlani3d_pipeline}
\end{figure*}
\chn{caption太长了。需要缩短到2行以内，大多数内容在正文描述清楚，节省空间}

\textbf{Prior Loss.} For images with only 2D annotations, to ensure that the predicted shape and pose parameters are reasonable, we enforce them to be close to a prior distribution by calculating the Mahalanobis distance. The prior loss is
\begin{equation}
    \label{prior_loss}
    \mathcal{L}_{\text{prior}} = \lambda_{\beta} (\hat{\beta} - \mu_{\beta})^{T} \Sigma^{-1}_{\beta} (\hat{\beta} - \mu_{\beta}) + (\hat{\theta} - \mu_{\theta})^{T} \Sigma^{-1}_{\theta} (\hat{\theta} - \mu_{\theta}),
\end{equation}
where \(\lambda_{\beta}=0.5\), \(\mu_{\beta}\), \(\Sigma_{\beta}\), \(\mu_{\theta}\), and \(\Sigma_{\theta}\) are the mean and covariance of the prior distributions given by SMAL~\citep{zuffi20173d}.

\textbf{Adversarial Loss.} Finally, we adopt a discriminator to further ensure that the model outputs natural poses and shapes, therefore we employ an adversarial loss similar to HMR~\citep{kanazawa2018end}. This loss is designed to make the predicted parameters indistinguishable from real distribution:
\begin{equation}
    \label{adversarial_loss}
    \mathcal{L}_{\text{adv}} = \sum_{k} (D_{k}(\theta, \beta) - 1)^{2},
\end{equation}
where \(D_{k}\) represents a discriminator. 

\section{CtrlAni3D Dataset}
\label{sec:ctrlani3d}



\textbf{SMAL Structure Condition.} The dataset generation pipeline is illustrated in Fig.~\ref{fig:ctrlani3d_pipeline}. Given a posed SMAL mesh and viewpoint, we render it into mask map and depth map as the condition images to guide the structure of images generated by a pre-trained ControlNet~\citep{zhang2023adding}. To guarantee the diversity of the poses and shapes, we randomly sample \(\beta\) from the Gaussian distributed shape space provided by SMAL~\citep{zuffi20173d}, and sample a more diverse range of \(\theta\) from a combined pose space presented by dog~\citep{ruegg2023bite, biggs2020wldo} and horse motion~\citep{li2024poses}. This is reasonable because the quadrupeds expressed by SMAL share similar anatomical structures. About the viewpoint for rendering, each dimension of the global rotation vector is uniformly sampled from $(-\pi,\pi)$ while the position is uniformly sampled between $[-0.5,-0.5,4]$ and $[0.5,0.5,8]$, therefore possible truncated images may be generated to enhance the robustness of training.

\textbf{Text Condition.} To further control the style of generated images $x$, we seek to use text prompt. We first manually classify the sampled 3D meshes into 10 species: cat, tiger, lion, cheetah, dog, wolf, horse, zebra, cow and hippo, and use species name as one keyword. When hard to distinguish (e.g., some hard cases of \textbf{zebra/horse}), we use both animal names as prompts and select the one with the highest cycle consistency. The second keyword is the pose description, e.g. "stands" in Fig.~\ref{fig:ctrlani3d_pipeline}, which is assigned by human annotator according to the 3D mesh. Based on these keywords, ChatGPT~\citep{achiam2023gpt} is employed to complete a prompt sentence $c_t$ depicting possible animal behaviors. Finally, both $c_v$ and $c_t$ act as the prompts of ControlNet for realistic and rich image generation.

\textbf{Semi-Automated Filtering.} Note that, not all generated images are perfectly aligned with conditions. To address this, we design a semi-automated filtering strategy to lower the burden of annotators. First, SAM2~\citep{ravi2024sam2} is utilized to extract the foreground mask of the generated images, enabling cycle-consistency checking by comparing to conditioned mask. 
Further, we manually filter out images that do not match the mesh poses to ensure good data quality. 
Each synthetic image has a resolution of \(512 \times 512 \) and includes well-aligned annotations 
for \(\beta\), \(\theta\), \(\gamma\) and 3D keypoints. By comparing rendered depth image the projected keypoint depths, we also obtain visible 2D keypoints as annotation. 
Supplementary Material demonstrates both success cases and failure cases that are manually filtered. 

\textbf{Backgrounds.}
To ensure the background diversity, we randomly sample COCO~\citep{lin2014microsoft} images to serve as backgrounds for \(2/7\) of the CtrlAni3D images. The selection of the proportion of the COCO background can be referenced in the Supplementary Materials.
\section{Experiments}

\subsection{Setup}
\textbf{Datasets.}
We curate and aggregate multiple datasets containing 2D and 3D annotations for animals. The full dataset include the training split of Animal Pose~\citep{cao2019cross}, APT-36K~\citep{yang2022apt}, AwA-Pose~\citep{banik2021novel}, Stanford Extra~\citep{biggs2020wldo}, Zebra synthetic~\citep{zuffi2019three}, Animal3D~\citep{xu2023animal3d}, and our own CtrlAni3D. Only Animal3D, CtrlAni3D and Zebra synthetic have 3D annotations. The full dataset contains 41.3k images, and validation ratio is \(3/20\). For evaluation, we mainly use the test part of Animal3D and CtrlAni3D, and Animal Kingdom dataset~\citep{Ng_2022_CVPR}. Note that, Animal Kingdom serves as out-of-distribution (OOD) benchmark because it contains very challenging in-the-wild images which are never seen during training. Only 8 quadruped species from Animal Kingdom are selected for testing. 
For more details about the datasets, please refer to the Supplementary Material. 

\noindent\textbf{Training Details.}
Our model is implemented by Pytorch. We use AdamW~\citep{adamw} optimizer with a linear learning rate decay schedule. The initial learning rate is $1.25 \times 10^{-6}$. The entire training takes a week on a NVIDIA RTX 4090 GPU.

\noindent\textbf{Metrics.} Several 3D and 2D metrics are employed to fully assess the model performance, listed below. \textit{PA-MPJPE} is Procrustes-Aligned Mean Per Joint Position Error for regressed 3D keypoints. 
\textit{PA-MPVPE} is Procrustes-Aligned Mean Per Vertex Position Error for SMAL vertices. 
\textit{PCK} is Percentage of Correct Keypoints given a threshold. In this paper, PCK is only used for evaluating 2D keypoints. PCK@HTH uses half the head-to-tail distance as the threshold. By setting threshold to 0.1 and 0.15, we get commonly used PCK@0.1 and PCK@0.15 metrics. 
\textit{AUC} is Area Under the Curve value when the PCK threshold gradually increase from 0 to 1. Note that, MPJPE without Procrustes-Alignment is not used because large animal size variations would cause imbalanced MPJPE among species.

\begin{table*}[bp]
\caption{\textbf{Quantitative comparisons on Animal3D, CtrlAni3D and AnimalKingdom datasets.} 
\textbf{Bold} numbers indicate the best values. P@H, P@0.1, P@0.15, PAJ, and PAV represent PCK@HTH, PCK@0.1, PCK@0.15, PA-MPJPE, and PA-MPVPE, respectively. 
}
\label{table:comparison results to baseline}
\centering
\resizebox{0.9\textwidth}{!}{
\begin{tabular}{ccccccccccccc}
\toprule
Dataset & \multicolumn{4}{c}{Animal3D} & \multicolumn{4}{c}{CtrlAni3D} & \multicolumn{4}{c}{Animal Kingdom} \\ 
\cmidrule(lr){2-5} \cmidrule(lr){6-9} \cmidrule(lr){10-13} 

Metric  & AUC\(\uparrow\) & P@H\(\uparrow\) & PAJ\(\downarrow\) & PAV\(\downarrow\) & AUC\(\uparrow\) & P@H\(\uparrow\) & PAJ\(\downarrow\) & PAV\(\downarrow\) & AUC\(\uparrow\) & P@H\(\uparrow\) & P@0.1\(\uparrow\) & P@0.15\(\uparrow\) \\ \hline

HMR    & 76.3 & 60.8 & 123.5 & 133.9 & 80.8 & 67.0 & 123.5 & 133.9 & 70.2 & 64.0 & 12.8 & 25.6 \\
WLDO   & 78.2 & 68.7 & 112.3 & 125.2 & 88.7 & 86.7 & 71.5 & 83.4 & 70.1 & 64.3 & 14.6 & 27.6 \\
AniMer-a & 75.2 & 57.2 & 115.5 & 128.7 & 80.3 & 66.0 & 117.0 & 129.4 & 68.9 & 62.5 & 10.2 & 21.3 \\
AniMer-b & 60.6 & 38.9 & 147.9 & 157.6 & 78.5 & 65.9 & 102.3 & 112.6 & 45.4 & 31.8 & 4.0 & 9.2 \\
HMR2.0 & 86.7 & 84.6 & 94.1 & 98.5 & 91.8 & 93.0 & 60.9 & 66.4 & 77.3 & 73.9 & 22.7 & 40.2 \\
\midrule
AniMer  & \textbf{88.9} & \textbf{89.5} & \textbf{80.4} & \textbf{85.7} & \textbf{93.8} & \textbf{95.4} & \textbf{44.1} & \textbf{47.6} & \textbf{82.9} & \textbf{83.7} & \textbf{34.9} & \textbf{54.7} \\ \bottomrule
\end{tabular}
}

\end{table*}



\begin{figure*}[ht]
    \centering
     \includegraphics[width=0.85\linewidth]{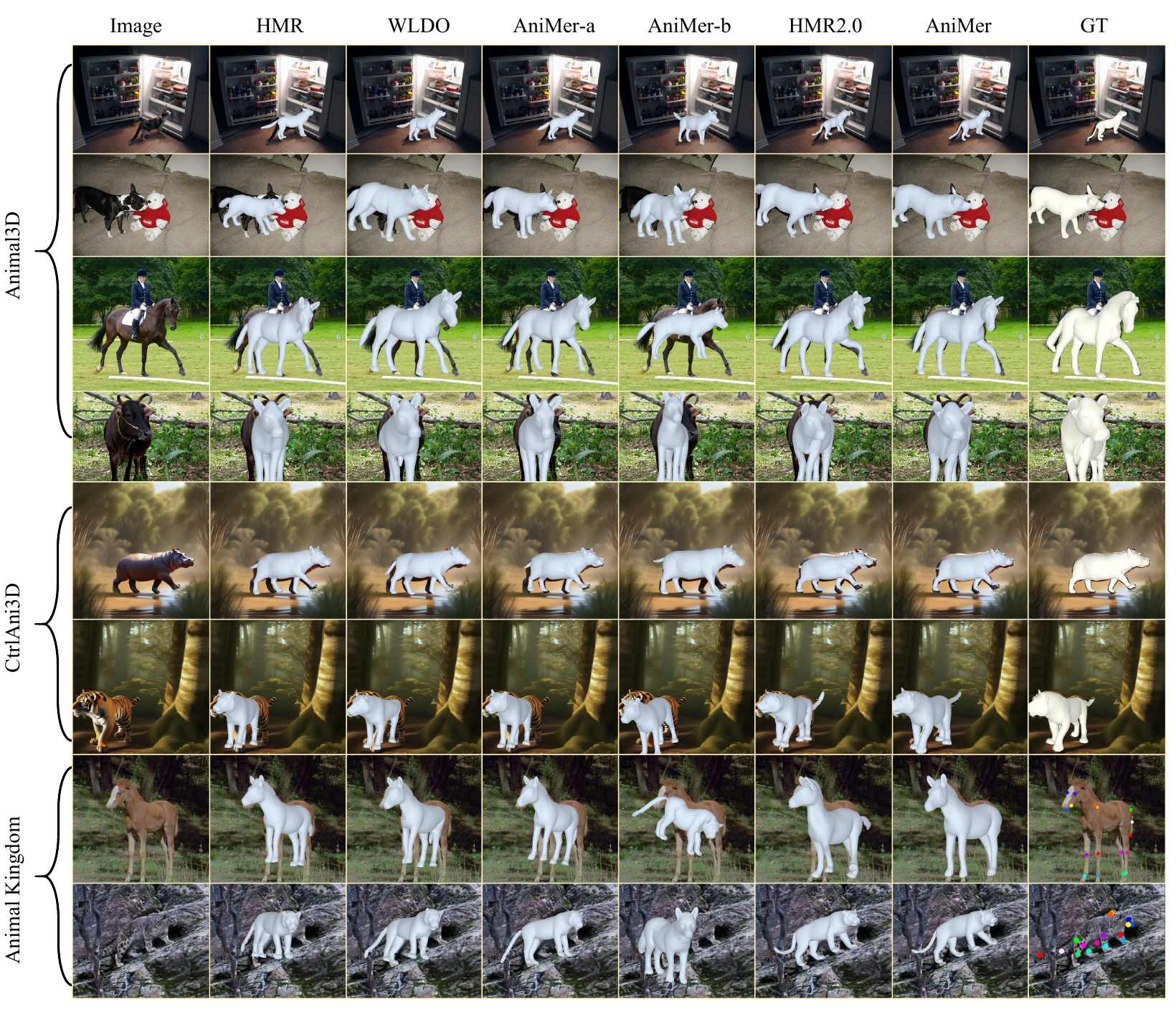}
    \caption{\textbf{Qualitative comparisons on Animal3D, CtrlAni3D and Animal Kingdom datasets.} We compare our results with HMR~\citep{kanazawa2018end}, WLDO~\citep{biggs2020wldo}, AniMer-a (ResNet152 backbone), AniMer-b (no pretraining) and HMR2.0~\citep{goel2023humans}. 
    }
    \label{fig:quali_compare_main}
\end{figure*}

\subsection{Multi-Species Experiments}
Following the practice of Animal3D dataset~\citep{xu2023animal3d}, we utilize HMR~\citep{kanazawa2018end} and WLDO~\citep{biggs2020wldo} as baselines. 
Furthermore, to highlight the significance of our ViT backbone and backbone pretraining, we compare the final AniMer with two variants: AniMer-a which replaces ViT backbone with ResNet152, and AniMer-b which discards ViT pretraining. To be fair, all above methods are trained on our full dataset with the same losses and the same two-stage training strategies. Also, we compare to HMR2.0~\citep{goel2023humans} to highlight our special design choices. Ablation study on different encoder and decoder designs can be found in the Supplementary Material.

Quantitative results are shown in Tab.~\ref{table:comparison results to baseline}. We observed that our AniMer method achieved state-of-the-art results and that both 3D metrics and 2D metrics consistently outperformed previous work on multiple datasets.
Specifically on Animal Kingdom dataset, AniMer demonstrates strong robustness and improves over the previous work for all the metrics. 
Qualitative results in Fig.~\ref{fig:quali_compare_main} show that AniMer aligns with images much better for thin structures such as legs and tails. More qualitative results are shown in Supplementary Material.

\subsection{Ablation Study on CtrlAni3D}


To further demonstrate the effectiveness of our proposed CtrlAni3D dataset in enhancing network capabilities, we conduct a series of ablation experiments. 
We compare models trained on different combinations of Animal3D dataset (A3D), CtrlAni3D dataset (C3D), and other datasets (others). The hyperparameter training settings remain unchanged.

\begin{table}[ht]
\centering
\begin{small}
\caption{\textbf{Effect of CtrlAni3D on 2D datasets.} 
We report AUC, PCK@0.1 and PCK@0.15 as 2D metrics. \textbf{Bold} numbers indicate the best performance.}
\label{tab:ablation_2D}
\resizebox{\columnwidth}{!}{%
\begin{tabular}{c@{\hskip 2.0pt}c@{\hskip 2.0pt}c@{\hskip 2.0pt}c@{\hskip 2.0pt}c@{\hskip 2.0pt}c@{\hskip 2.0pt}c}
\toprule
\multicolumn{1}{c}{} & \multicolumn{3}{c}{Training Data} & \multicolumn{3}{c}{2D Metric} \\
\cmidrule(lr){2-4} \cmidrule(lr){5-7} 
\makecell{Testset} & \multicolumn{1}{c}{A3D}& \multicolumn{1}{c}{C3D}& \multicolumn{1}{c}{others}& \multicolumn{1}{c}{AUC$\uparrow$}& 
\multicolumn{1}{c}{PCK@0.1$\uparrow$}& \multicolumn{1}{c}{PCK@0.15$\uparrow$} \\

\midrule
\multirow{4}{*}{\makecell{Animal\\Kingdom}
} 
&     &\dui &\dui &
82.7  & 33.8 & 53.2 \\
 &  \dui & \dui &      &
81.9  & 32.1 & 51.8 \\
 &\dui &      & \dui &
82.1 & 33.7 & 53.2 \\
 &  \dui & \dui & \dui &
\textbf{82.9} & \textbf{34.9} & \textbf{54.7} \\

\midrule
\multirow{4}{*}{ \makecell{Animal \\Pose} } 
&     &\dui &\dui &
85.5  & 41.5 & 63.7 \\
 &  \dui & \dui &      &
82.9  & 33.4 & 55.4 \\
 &\dui &      & \dui &
85.3  & 40.4 & 63.0 \\
 &  \dui & \dui & \dui &
\textbf{85.8} & \textbf{42.9} & \textbf{65.1} \\

\bottomrule
\end{tabular}
}
\end{small}
\end{table}

\begin{figure}[ht]
    \centering
    \includegraphics[width=1.0\linewidth]{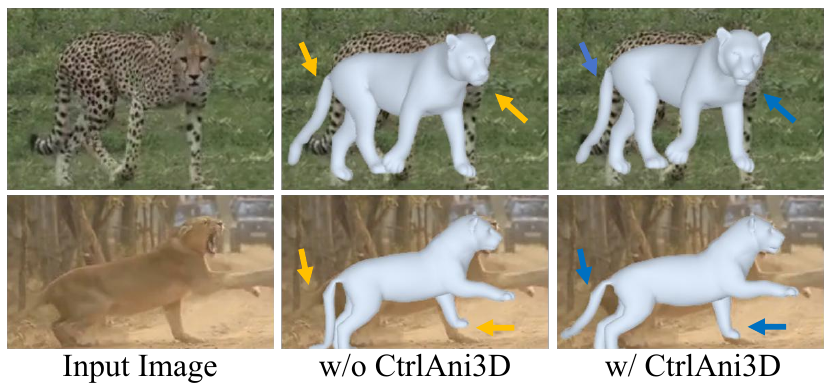}
    \caption{\textbf{Effect of CtrlAni3D on Animal Kingdom dataset.} Input and result images are zoom-in cropped for visualization. Training with CtrlAni3D enhances the model's ability to align tails, limbs and faces. {\color{orange}Orange} arrows indicate misalignments, while {\color{blue}blue} arrows indicate better alignments. }
    \label{fig:without_ctrlani3d}
\end{figure}

Specifically, Tab.~\ref{tab:ablation_2D} indicates that our model, when trained with CtrlAni3D, exhibits superior performance even on previously unseen in-the-wild data. Qualitative comparisons in 
Fig.~\ref{fig:without_ctrlani3d} further demonstrates that training with CtrlAni3D helps AniMer to yield more accurate animal terminal body parts such as tails, limbs and faces. 

To further emphasize the superiority of our diffusion based pipeline over traditional CG synthetic dataset, we follow the experiments performed by Animal3D~\citep{xu2023animal3d}, see Tab.~\ref{tab:synthe}. Specifically, ``HMR'' means directly training HMR on Animal3D. ``HMR-Synthetic'' pretrains HMR with a CG-based synthetic dataset generated by textured SMAL~\citep{zuffi2018lions} for 100 epochs before training on Animal3D. Similarly, ``HMR-CtrlAni3D" pretrains HMR on CtrlAni3D for 100 epochs. It is clearly shown that CtrlAni3D provides better pretraining than traditional CG based synthetic data. Additional ablation studies on CtrlAni3D, as well as a comparison between CtrlAni3D and Animal3D, can be found in the Supplementary Material. 


\begin{table}[ht]
\centering
\caption{\textbf{CtrlAni3D v.s. CG synthetic data.} The values of ``HMR-Synthetic'' are borrowed from paper ~\citet{xu2023animal3d}.}
\label{tab:synthe}
\resizebox{0.75\columnwidth}{!}{
\begin{tabular}{ccc}
\toprule
& \multicolumn{2}{c}{Animal3D}                                 \\ \cline{2-3} 
\multirow{-2}{*}{Method}          & {PCK@HTH\(\uparrow\)}  & {PA-MPJPE\(\downarrow\)}                      \\ \hline
{HMR}                                &60.5 & 127.8 \\
{HMR-Synthetic}          &63.1 & 124.8 \\
{HMR-CtrlAni3D}  &\textbf{64.0} & \textbf{121.9} 
\\ \bottomrule
\end{tabular}
}
\end{table}

\subsection{Ablation Study on Animal Family Supervised Contrastive Learning}
\begin{table}[ht]
\caption{\textbf{Effect of animal family supervised contrastive learning.} ``*'' indicates training on Animal3D and CtrlAni3D only. PAJ: PA-MPJPE. PAV: PA-MPVPE. P@0.1: PCK@0.1. }
\label{tab: Effective of Animal family supervised contrastive learning}
\resizebox{\columnwidth}{!}{%
\begin{tabular}{ccccccc}
\toprule
\multirow{2}{*}{Method} & \multicolumn{2}{c}{Animal3D} & \multicolumn{2}{c}{CtrlAni3D} & \multicolumn{2}{c}{Animal Kingdom} \\ \cmidrule(lr){2-3} \cmidrule(lr){4-5} \cmidrule(lr){6-7}
      & PAJ\(\downarrow\) & PAV\(\downarrow\) & PAJ\(\downarrow\) & PAV\(\downarrow\) & AUC\(\uparrow\) & P@0.1\(\uparrow\) \\ \hline
w/o \(\mathcal{L}_{\text{con}}\)* 
& 82.5 & 88.0 & 54.6 & 59.2 & 81.4 & 30.4    \\ 
w/ \(\mathcal{L}_{\text{con}}\)* 
& \textbf{80.8} & \textbf{86.0} & \textbf{46.1} & \textbf{50.2} & \textbf{81.9} & \textbf{32.1}    \\ 
w/o \(\mathcal{L}_{\text{con}}\) 
& 81.3 & 86.7 & 44.7 & 48.4 & 82.7 & 34.4    \\ 
w/ \(\mathcal{L}_{\text{con}}\)  
& \textbf{80.4} & \textbf{85.7} & \textbf{44.1} & \textbf{47.6} & \textbf{82.9} & \textbf{34.9}   \\ 
\bottomrule
\end{tabular}%
}
\end{table}

Finally, we conduct experiments to demonstrate the effects of animal family supervised contrastive learning. Tab.~\ref{tab: Effective of Animal family supervised contrastive learning} presents the quantitative results on Animal3D, CtrlAni3D, and Animal Kingdom. As we can see, AniMer trained with \(\mathcal{L}_{\text{con}}\) outperforms the model that does not utilize \(\mathcal{L}_{\text{con}}\). The improved PA-MPVPE and PA-MPJPE metrics indicate that \(\mathcal{L}_{\text{con}}\) enhances the model's capabilities in shape and pose estimation, respectively.

\begin{figure}[ht]
    \centering
    \includegraphics[width=0.9\linewidth]{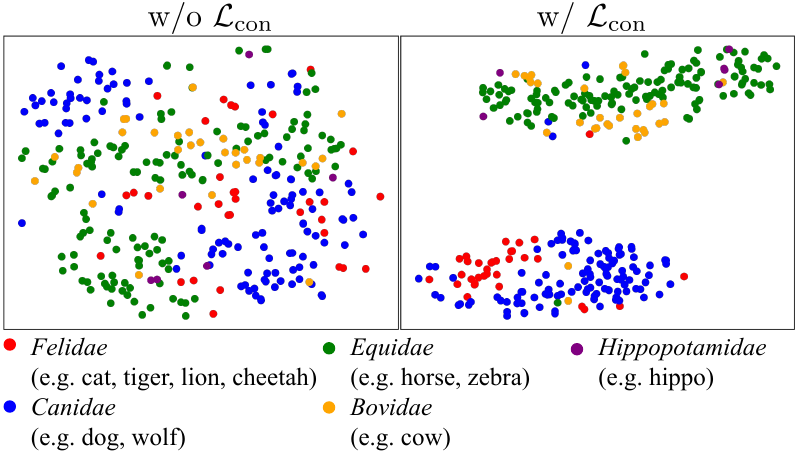}
    \caption{\textbf{The t-SNE~\citep{van2008visualizing} visualization of the class token on the Animal3D test set.} With the inclusion of \(\mathcal{L}_\text{con}\), features of animals from the same family are more closely clustered.}
    \label{fig:tsne_anisupcon}
\end{figure}

\begin{figure}[ht]
    \centering
    \includegraphics[width=1.0\linewidth]{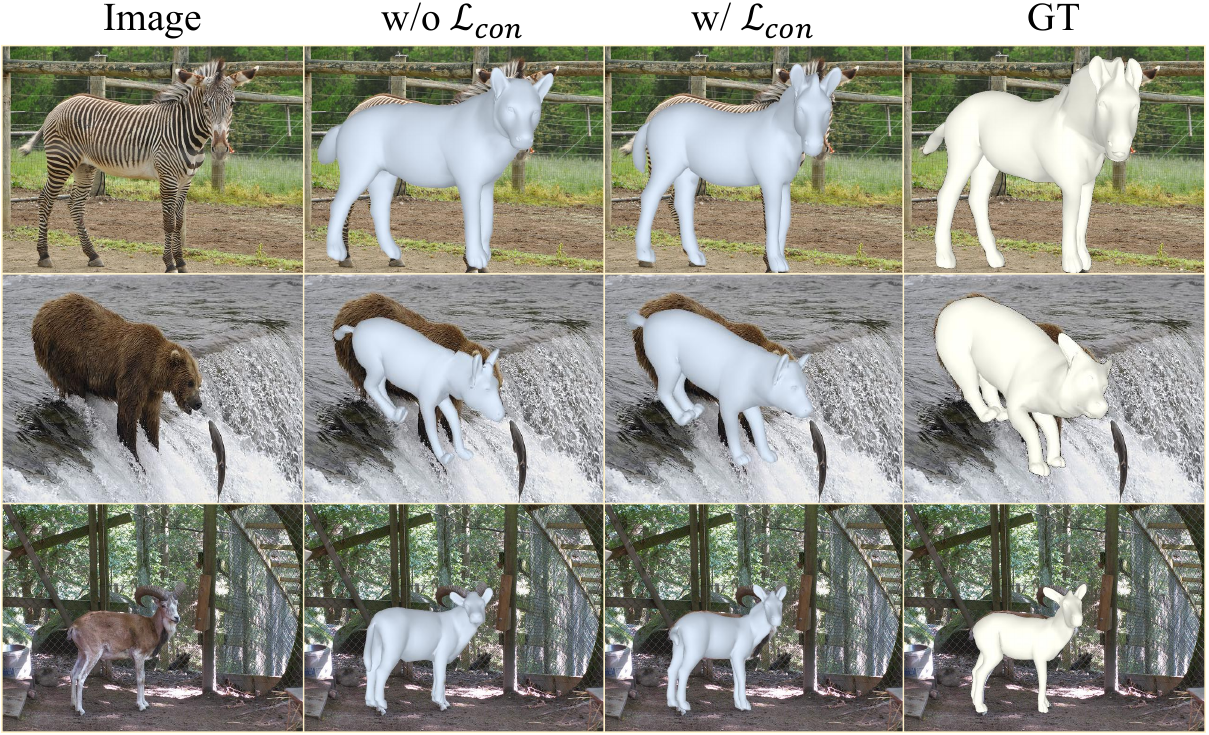} 
    \caption{\textbf{Effect of animal family supervised contrastive learning.} Without animal family supervised contrastive learning, the estimation is prone to incorrect shapes for animal families.} 
    \label{fig:wo_anisupcon}
\end{figure}

In addition, to provide a more intuitive understanding of the effect of 
\(\mathcal{L}_{\text{con}}\), Fig.~\ref{fig:tsne_anisupcon} presents the t-SNE~\citep{van2008visualizing} visualization of the class token on Animal3D test set. It is evident that \(\mathcal{L}_{\text{con}}\) brings the features of animals from the same family closer together, thereby enhancing the model's ability to distinguish between the shapes of animals from different families. Fig.~\ref{fig:wo_anisupcon} further demonstrates that \(\mathcal{L}_{\text{con}}\) enables the model to better distinguish between different animal families. As shown in Fig.~\ref{fig:wo_anisupcon}, when trained without \(\mathcal{L}_{\text{con}}\), the model incorrectly predicts a zebra, bear, and antelope as a dog, zebra, and cow, respectively. More discussion about family contrastive loss can be found in the Supplementary Material.

\section{Conclusion}
\textbf{Summary. }
This paper presents AniMer, a simple yet effective method for accurate animal pose and shape estimation. The key to the success of AniMer is a large capacity Transformer backbone together with an aggregated large scale dataset. Within the aggregated dataset, we propose a novel synthetic general quadruped dataset CtrlAni3D, which is rendered by prompting a controllable text-to-image generation model ControlNet. Benefiting from our family aware design philosophy, AniMer outperforms previous methods not only on 3D quadrupedal datasets, but also on 2D in-the-wild datasets. We believe the principles behind AniMer would inspire the mesh recovery tasks of the whole animal kingdom, and enable several down-stream applications such as avatar creation and behavioral analysis. 

\noindent\textbf{Limitations and Future Work.}
The solution space of AniMer is limited to the pose and shape space of SMAL model. For species (e.g. birds or goats) whose shapes are far from the linear shape space of SMAL, building specific animal models, enhancing vertex deformations and introducing bone length scale are necessary to yield pixel-aligned results. Besides, motion blur and severe occlusions would harm the reconstruction results. In the future, AniMer would be strengthened with better animal parametric models and spatio-temporal methods. 

\noindent\textbf{Acknowledgements.} This study was supported by the National Key Research and Development Program of China (2023YFC2415400); the National Natural Science Foundation of China (T2422012, 62071210, 62125107); the Guangdong Basic and Applied Basic Research (2024B1515020088); the Shenzhen Science and Technology Program (RCYX20210609103056042); the Guangdong Basic and Applied Basic Research (2021A1515220131); the High Level of Special Funds (G030230001, G03034K003); the Shuimu Tsinghua Scholar Program (2024SM324). 




{
    \small
    \bibliographystyle{ieeenat_fullname}
    \bibliography{reference}
}

\appendix
\clearpage
\setcounter{page}{1}
\maketitlesupplementary

\section{Datasets}
\textbf{Animal Pose dataset.} The Animal Pose dataset \citep{cao2019cross} includes five categories: dog, cat, cow, horse and sheep, comprising a total of over 6,000 instances across more than 4,000 images. Each animal instance in Animal Pose dataset is annotated with 20 keypoints.

\noindent\textbf{APT-36k dataset.} The APT-36k dataset \citep{yang2022apt} contains 36000 images covering 30 different animal species from different scenes. There are typically 17 keypoints labeled for each animal instance.

\noindent\textbf{AwA Pose dataset.} The AwA Pose dataset \citep{banik2021novel} is introduced for 2D quadruped animal pose estimation. AwA contains 10064 images of 35 quadruped animal species and each image is annotated with 39 keypoints.

\noindent\textbf{Stanford Extra dataset.} The Stanford Extra dataset \citep{biggs2020wldo} consists of 20,580 images and covers 120 dog breeds. Each image is annotated with 20 2D keypoints and silhouette.

\noindent\textbf{Zebra synthetic dataset.} The Zebra synthetic dataset \citep{zuffi2019three} consists of 12850 images. Each image is randomly generated that differs in background, shape, pose, camera, and appearance.

\noindent\textbf{Animal Kingdom dataset.} The Animal Kingdom dataset \citep{Ng_2022_CVPR} includes a diverse range of animal species. We only use 8 major animal classes of pose estimation dataset to evaluate our method. 

\noindent\textbf{Animal3D dataset.} Animal3D dataset \citep{xu2023animal3d} contains a total of 3379 images, which are classified into 40 classes. Each image is annotated with SMAL \citep{zuffi20173d} parameters, 2d keypoints, 3d keypoints and masks.

\noindent\textbf{CtrlAni3D dataset.} Our dataset is annotated in the same style as Animal3D dataset. More details about our dataset can be found in Sec.~\ref{sec: more details about ctrlani3d}.

For all datasets, we filter out images of animals not included in SMAL~\citep{zuffi20173d}, such as elephants. We then aggregate all the aforementioned datasets (excluding Animal Kingdom) for training, assigning different sampling weights to each dataset based on its type and size, as shown in Tab.~\ref{tab:train_data_weights}.
\begin{table}[ht]
\centering
\caption{\textbf{Full dataset statistics for training.}}
\resizebox{0.95\columnwidth}{!}
{
\begin{tabular}{lllc}
\toprule
Dataset         & Number & Ratio  & Training Sample Weight \\ \hline
Animal3D        & 3065   & 7.4\%  & 1                      \\
CtrlAni3D       & 8277   & 20.0\% & 0.5                    \\
Animal Pose     & 1680   & 4.0\%  & 0.15                   \\
AwA-Pose        & 2884   & 7.0\%  & 0.15                   \\
Zebra Synthetic & 12850  & 31.1\% & 0.05                   \\
Stanford Extra  & 7689   & 18.6\% & 0.15                   \\
APT-36K         & 4887   & 11.8\% & 0.15                   \\
Total           & 41332  & 100\%  & -                      \\ \bottomrule
\end{tabular}
}
\label{tab:train_data_weights}
\end{table}

\section{More Details about CtrlAni3D}
\label{sec: more details about ctrlani3d}
Each image in the CtrlAni3D dataset is annotated with SMAL parameters, including \(\beta \in \mathbb{R}^{41}\), \(\theta \in \mathbb{R}^{35 \times 3}\) (expressed by axis angle), and \(\gamma \in \mathbb{R}^{3}\). Additionally, similar to Animal3D~\citep{xu2023animal3d}, CtrlAni3D provides annotations for 26 3D keypoints and their corresponding 2D keypoints. The visibility of 2D keypoints is determined by comparing the depth \(d_k\) of the keypoint with the depth \(d_p\) at the corresponding pixel location. Specifically, visibility is set to 1 when \(d_k \le d_p\); otherwise, it is set to 0.

\begin{figure*}
    \centering
    \includegraphics[width=\linewidth]{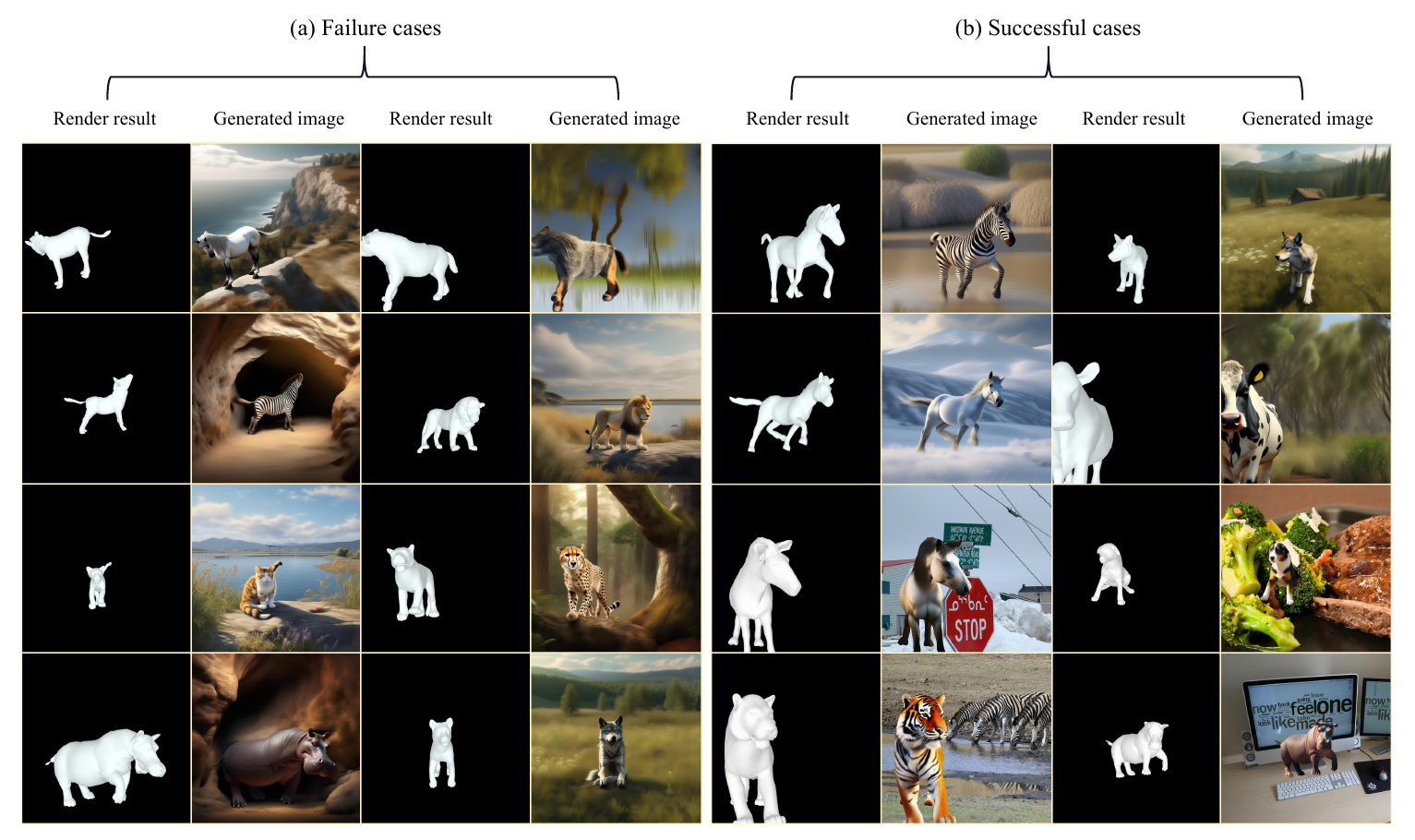}
    \caption{\textbf{CtrlAni3D failure cases and successful cases.} (a) Failure cases. There are two main cases of failure: (1) At times, ControlNet may struggle to generate mesh-aligned poses (first row and second row). (2) Additionally, ControlNet may not effectively generate the intricate details of the animal body (third row and fourth row). (b) Successful cases. The backgrounds of the first and second rows are generated by ControlNet, while the backgrounds of the third and fourth rows are sourced from the COCO dataset.}
    \label{fig:data_generate_failure_cases}
\end{figure*}

During the generation of CtrlAni3D dataset, we prompt the ControlNet using common names instead of scientific names of animals. However, to better indicate the position our CtrlAni3D dataset in the animal taxonomy, we list the most relevant scientific names of used animal species in Tab.~\ref{tab:taxonomy}. During the image generation process, we require human annotators to filter out misaligned results, as described in main text Sec.4. Such misaligned results, denoted as ``Failure cases'', are illustrated in Fig.~\ref{fig:data_generate_failure_cases}. 

In addition, COCO backgrounds are used only when SAM2 achieves satisfactory segmentation quality. Therefore, increasing the ratio of COCO backgrounds is equivalent to lowering the IoU threshold, which results in a decline in data quality. This may negatively impacts model training.

\begin{table}[ht]
\centering
\caption{\textbf{Scientific names of used animal species in CtrlAni3D. } The image counts of each species are listed at the right column. }
\resizebox{0.46\textwidth}{!}
{
\begin{tabular}{llll} \toprule
Family                   & Species                & Prompt Commands & Count \\ \hline
\multirow{4}{*}{
\textit{Felidae}} & \textit{Felis catus} & Cat & 80 \\
                  & \textit{Panthera leo} & Lion & 630 \\
                  & \textit{Acinonyx jubatus} & Cheetah & 299 \\
                  & \textit{Panthera tigris  } & Tiger & 280 \\ \hline
\multirow{2}{*}{
\textit{Canidae}} & \textit{Canis lupus familiaris} & Dog & 2976 \\
                  & \textit{Canis lupus } & Wolf & 413 \\ \hline
\multirow{2}{*}{
\textit{Equidae}} & \textit{Equus ferus caballus} & Horse & 2228 \\
                  & \textit{Equus zebra} & Zebra & 1460 \\ \hline
\textit{Bovidae}  & \textit{Bos taurus}  & Cow & 890 \\ \hline
\textit{Hippopotamidae} & \textit{Hippopotamus amphibius} & Hippo & 455 \\ \hline
\multicolumn{3}{l}{Total} & 9711   \\ \bottomrule
\end{tabular}
}
\label{tab:taxonomy}
\end{table}

\section{Comparison between CtrlAni3D and Animal3D}
\begin{figure}[ht]
    \centering
    \includegraphics[width=\columnwidth]{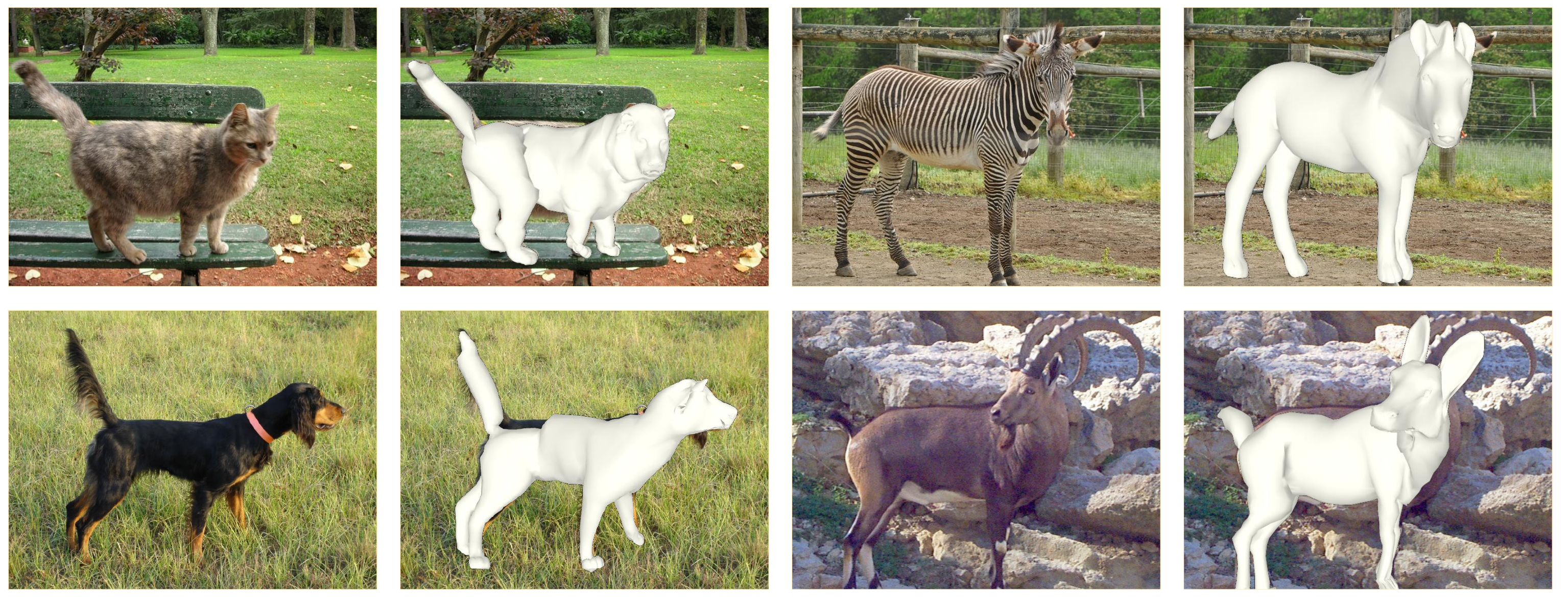}
    \caption{\textbf{Some bad cases in Animal3D.}}
    \label{fig:failure cases in Animal3D}
\end{figure}

CtrlAni3D and Animal3D are both based on SMAL. As a result, both datasets encompass five animal families, as presented in Tab.~\ref{tab:taxonomy}. Animal3D includes more subcategories (e.g., bighorn) compared to CtrlAni3D, which has a greater number of entries. However, there are some animal species that SMAL doesn't express very well (in the second row of Fig.~\ref{fig:failure cases in Animal3D}). In addition, Animal3D requires manual 2D annotations for fitting, which introduces a degree of error, as shown in the first row of Fig.~\ref{fig:failure cases in Animal3D}. CtrlAni3D ensures the quality of data through cycle consistency and manual filtering.

\section{More results and analysis.}
\textbf{Effect of CtrlAni3D for 3D pose estimation.} To demonstrate the improvements of CtrlAni3D for 3D pose estimation, we report the results of the 3D metrics on Animal3D, as shown in Tab.~\ref{tab: effect of including CtrlAni3D in training}. We can observe that training with CtrlAni3D enhances performance on Animal3D.

\begin{table}[ht]
\caption{\textbf{Effect of including CtrlAni3D in training.} We evaluate the performance of 3D pose estimation on two models. For the first model, we do not use CtrlAni3D during training, while for the second model, we incorporate CtrlAni3D into the training process.}
\centering
\resizebox{0.8\columnwidth}{!}{%
\label{tab: effect of including CtrlAni3D in training}
\begin{tabular}{ccc}
\toprule
                       & PA-MPJPE \(\downarrow\) & PA-MPVPE \(\downarrow\)  \\ \hline
AniMer(no CtrlAni3D)   & 82.6     & 88.4     \\ \hline
AniMer(with CtrlAni3D) & \textbf{80.4}    & \textbf{85.7}     \\ \bottomrule
\end{tabular}%
}
\end{table}

\noindent\textbf{Domain gap between CtrlAni3D and real-world data.} Although CtrlAni3D can improve the model performance on the real-world data, there still exist a certain domain gap. Tab.~\ref{tab: domain gap of ctrlani3d dataset} shows a certain domain gap between Animal3D and CtrlAni3D. However, the comparable results between AniMer(A3D) and AniMer(C3D) on the Animal Kingdom dataset indicate a similar generalization ability on in-the-wild data between Animal3D and CtrlAni3D. This is why we aggregate many datasets for full training. The performance gain shown in Tab.2 (in main text) further validates the effectiveness of CtrlAni3D to assist in generalization.
\begin{table}[ht]
\caption{\textbf{The generalizability of CtrlAni3D.} AniMer(A3D) trains only on Animal3D, AniMer(C3D) trains only on CtrlAni3D.}
\resizebox{\columnwidth}{!}{%
\label{tab: domain gap of ctrlani3d dataset}
\begin{tabular}{ccccccc}
\toprule
\multirow{2}{*}{Method} & \multicolumn{2}{c}{Animal3D} & \multicolumn{2}{c}{CtrlAni3D} & \multicolumn{2}{c}{Animal Kingdom} \\ \cline{2-7} 
                 & PCK@HTH \(\uparrow\) & AUC \(\uparrow\)  & PCK@HTH \(\uparrow\) & AUC \(\uparrow\)  & PCK@HTH \(\uparrow\) & AUC \(\uparrow\)  \\ \hline
AniMer(A3D)  & 87.0    & 86.0 & 89.7    & 89.9 & 78.0    & 78.6 \\
AniMer(C3D) & 83.8    & 81.9 & 93.5    & 95.0 & 77.8    & 80.3 \\ \bottomrule
\end{tabular}%
}
\end{table}

\noindent\textbf{Ablations on different encoder and decoder.} To emphasize the significance of the ViT encoder and the SMAL transformer decoder, we substitute the ViT encoder with a ResNet-152 encoder and the SMAL transformer decoder with an MLP decoder, respectively. The results are presented in Tab.~\ref{tab: ablation on different encoder and decoder}.
\begin{table}[ht]
\caption{\textbf{Ablations on different encoder and decoder.} AniMer-b: use ResNet-152 as encoder. AniMer-e: use MLP as decoder.}
\resizebox{\columnwidth}{!}{%
\label{tab: ablation on different encoder and decoder}
\begin{tabular}{ccccccc}
\toprule
\multirow{2}{*}{Method} & \multicolumn{2}{c}{Animal3D}  & \multicolumn{2}{c}{CtrlAni3D} & \multicolumn{2}{c}{Animal Kingdom} \\ \cline{2-7} 
         & PA-J \(\downarrow\) & PA-V \(\downarrow\) & PA-J \(\downarrow\) & PA-V \(\downarrow\) & AUC \(\uparrow\)  & PCK@0.1 \(\uparrow\) \\ \hline
AniMer-b & 115.5 & 128.7 & 117.0 & 129.4 & 68.9 & 10.2  \\
AniMer-e & 83.9 & 89.2 & 55.8 & 60.9 & 81.9 & 31.6  \\
AniMer  & \textbf{80.4} & \textbf{85.7} & \textbf{44.1} & \textbf{47.6} & \textbf{82.9}    & \textbf{34.9}   \\ \bottomrule
\end{tabular}%
}
\end{table}

\noindent\textbf{More discussion about contrastive learning.} The contrastive loss directly impacts the feature tokens \( \mathbf{F} \), which in turn indirectly impacts the feature vectors \( \boldsymbol{f} \) and aligns features to model the global structure, capturing family differences. This ensures that the final output shape aligns more closely with the category of the input image. Compared with contrastive learning, \(\mathcal{L}_{\text{cls}}\) (``w $\mathcal{L}_{\text{cls}}$'' in Tab.~\ref{tab: The impact of the contrastive learning}) focuses solely on optimizing classification accuracy, which may not necessarily improve geometric parameter regression. Moreover, contrastive learning facilitates a more compact intra-class distribution and a more separable inter-class distribution in the feature space~\citep{khosla2020supervised}, thereby enhancing the model's capability for few-shot learning. In Tab.~\ref{tab: The impact of the contrastive learning}, we report the results for various animals. $\mathcal{L}_{\text{con}}$ can improves performance for animals with limited training samples (e.g., boars are less than one percent of the training set).


\begin{table}[ht]
\caption{\textbf{Evaluation of some species in A3D.} PA-J: PA-MPJPE, PA-V: PA-MPVPE.}
\resizebox{\columnwidth}{!}{%
\label{tab: The impact of the contrastive learning}
\begin{tabular}{ccccccc}
\toprule
\multirow{2}{*}{species} & \multicolumn{2}{c}{w \(\mathcal{L}_{\text{cls}}\)} & \multicolumn{2}{c}{w/o \(\mathcal{L}_{\text{con}}\)} & \multicolumn{2}{c}{w \(\mathcal{L}_{\text{con}}\)} \\ \cline{2-7} 
      & PA-J  & PA-V          & PA-J          & PA-V  & PA-J           & PA-V           \\ \hline
dog   & 74.9  & 81.3          & 72.1          & 76.7  & \textbf{71.1}  & \textbf{75.1}  \\
zebra & 66.3  & 68.3          & 60.4          & 63.7  & \textbf{60.6}  & \textbf{62.7}  \\
horse & 78.5  & 86.6          & 77.4          & 86.5  & \textbf{75.9}  & \textbf{84.1}  \\
cat   & \textbf{129.4} & \textbf{132.0}         & 134.5         & 136.1 & 131.2          & 132.8          \\
cow   & 83.0  & 86.0          & 80.3          & 84.8  & \textbf{78.1}  & \textbf{83.2}  \\
sheep & 83.9  & \textbf{88.0} & 83.8          & 91.1  & \textbf{80.1}  & 88.5           \\
bear  & 79.4  & 80.0          & \textbf{76.5} & 80.5  & 76.8           & \textbf{79.3}  \\
boar  & 126.5 & 158.7         & 119.1         & 150.5 & \textbf{115.9} & \textbf{142.6} \\ \bottomrule
\end{tabular}%
}
\end{table}

\begin{figure*}[ht]
    \centering
    \includegraphics[width=\linewidth]{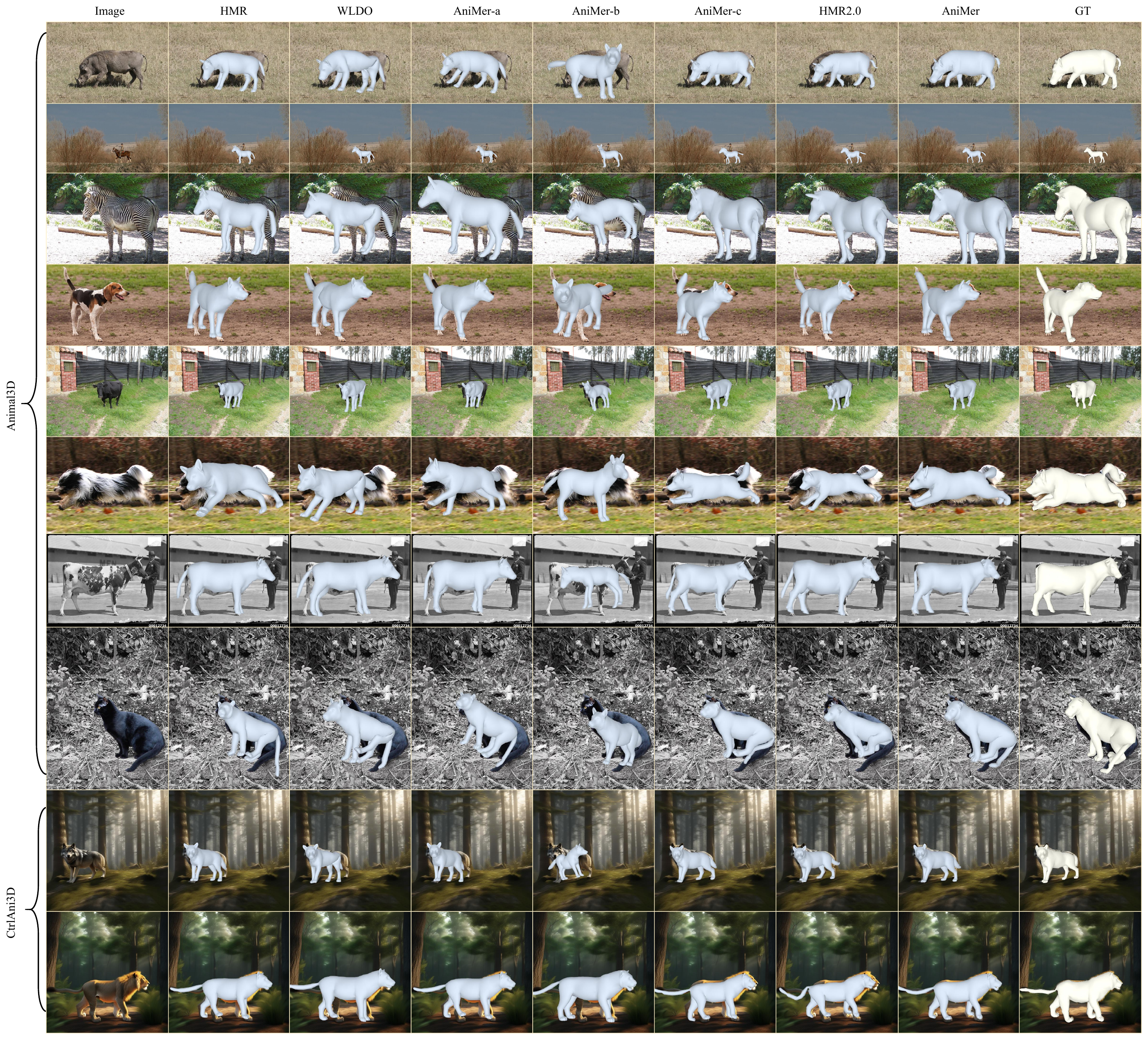}
    \caption{\textbf{More qualitative results on Animal3D and CtrlAni3D dataset.} We compare our results with HMR~\citep{kanazawa2018end}, WLDO~\citep{biggs2020wldo}, AniMer-a (ResNet152 backbone), AniMer-b (no pretraining), AniMer-c(train only one stage) and HMR2.0~\citep{goel2023humans}.}
    \label{fig:more qualititive results}
\end{figure*}

\noindent\textbf{Ablation study on two stage training.}
We compare to ``AniMer-c'', which trains the AniMer model for one stage. Both models are trained using the same batchsize and training steps. The two-stage training makes the model training more stable. By increasing the training steps or tuning the other hyperparameters, one-stage training may achieve comparable results. Quantitative results are shown in Tab.~\ref{tab:two differ from HMR2.0}, and qualitative results are shown in Fig.~\ref{fig:more qualititive results}.

\noindent\textbf{The effect of different setting.} Similar to the findings of \citep{goel2023humans}, our model exhibits varying performance metrics (PA-MPJPE(\(\downarrow\)) on Animal3D: \(87 - 78\), PCK@0.15\((\uparrow\)) on Animal Kingdom: \(0.5 - 0.6\)) under different settings (e.g., varying hyperparameters, different devices). 

\begin{table*}[ht]
\caption{\textbf{Quantitative comparisons on Animal3D, CtrlAni3D and AnimalKingdom datasets.} 
\textbf{Bold} numbers indicate the best values. P@H, P@0.1, P@0.15, PAJ, and PAV represent PCK@HTH, PCK@0.1, PCK@0.15, PA-MPJPE, and PA-MPVPE, respectively. 
}
\label{tab:two differ from HMR2.0}
\centering
\resizebox{0.9\textwidth}{!}{
\begin{tabular}{ccccccccccccc}
\toprule
Dataset & \multicolumn{4}{c}{Animal3D} & \multicolumn{4}{c}{CtrlAni3D} & \multicolumn{4}{c}{Animal Kingdom} \\ 
\cmidrule(lr){2-5} \cmidrule(lr){6-9} \cmidrule(lr){10-13} 

Metric  & AUC\(\uparrow\) & P@H\(\uparrow\) & PAJ\(\downarrow\) & PAV\(\downarrow\) & AUC\(\uparrow\) & P@H\(\uparrow\) & PAJ\(\downarrow\) & PAV\(\downarrow\) & AUC\(\uparrow\) & P@H\(\uparrow\) & P@0.1\(\uparrow\) & P@0.15\(\uparrow\) \\ \hline
AniMer-c & 87.2 & 86.3 & 85.9 & 90.4 & 91.7 & 93.4 & 59.5 & 64.2 & 80.6 & 80.4 & 28.6 & 47.5 \\
AniMer  & \textbf{88.9} & \textbf{89.5} & \textbf{80.4} & \textbf{85.7} & \textbf{93.8} & \textbf{95.4} & \textbf{44.1} & \textbf{47.6} & \textbf{82.9} & \textbf{83.7} & \textbf{34.9} & \textbf{54.7} \\ \bottomrule
\end{tabular}
}
\end{table*}

\section{More Qualitative Results}
\begin{figure*}[ht]
    \centering
    \includegraphics[width=0.85\linewidth]{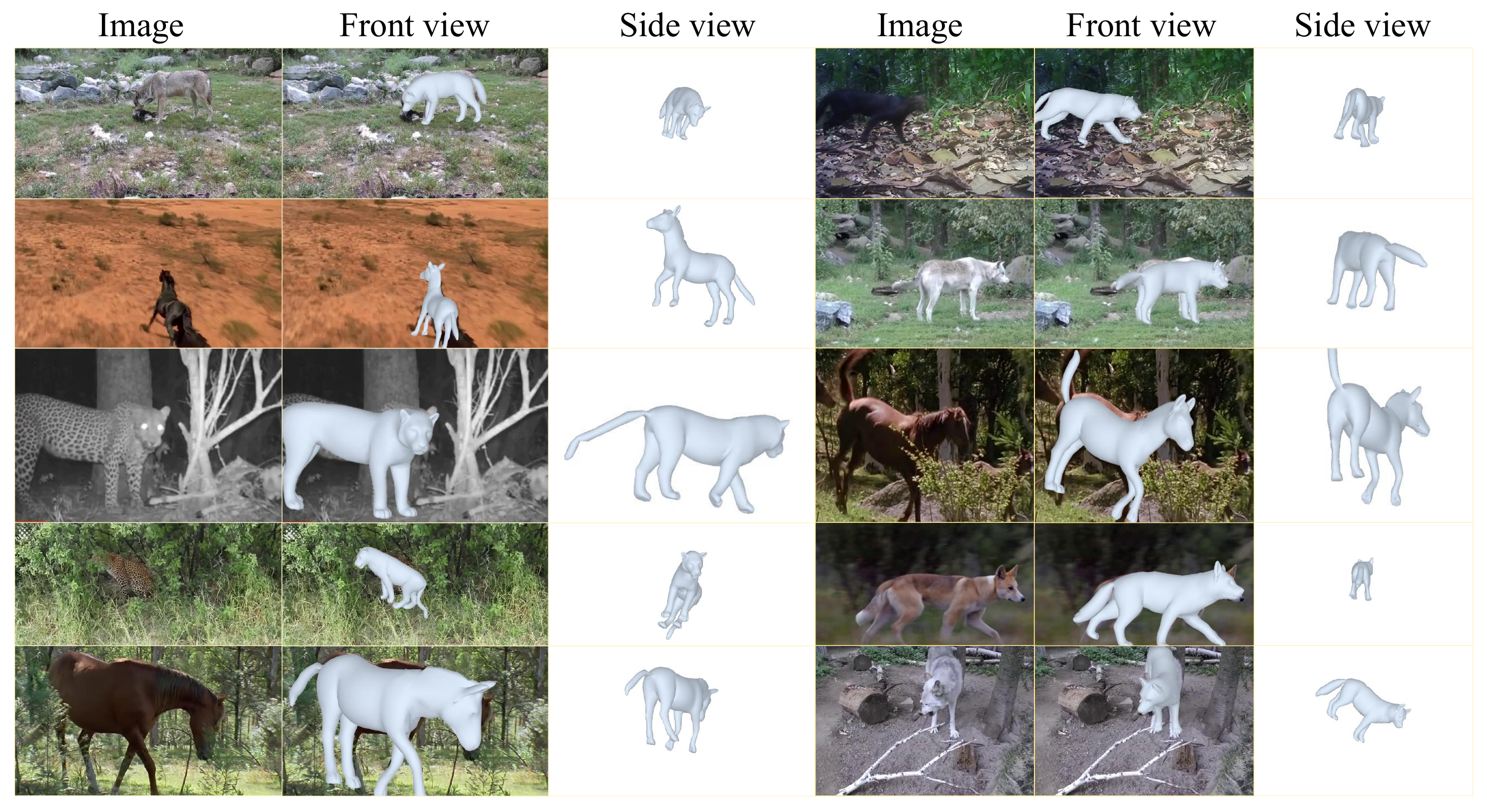}
    \caption{\textbf{Results on the Animal Kingdom dataset.} }
    \label{fig:qualitative results of animal kingdom}
\end{figure*}

We provide qualitative results from the Animal Kingdom dataset in Fig.~\ref{fig:qualitative results of animal kingdom}. For each case, we display the input image and the output results, which include both a front view rendering and a side view rendering. It can be observed that AniMer performs well even in challenging conditions such as motion blur (the second sample in the first row), unusual lighting (the first sample in the third row), partial occlusion (the first sample in the fourth row), and truncation (the second sample in the fifth row).

\section{Failure Cases and Discussion}
\begin{figure*}[ht]
    \centering
    \includegraphics[width=\columnwidth]{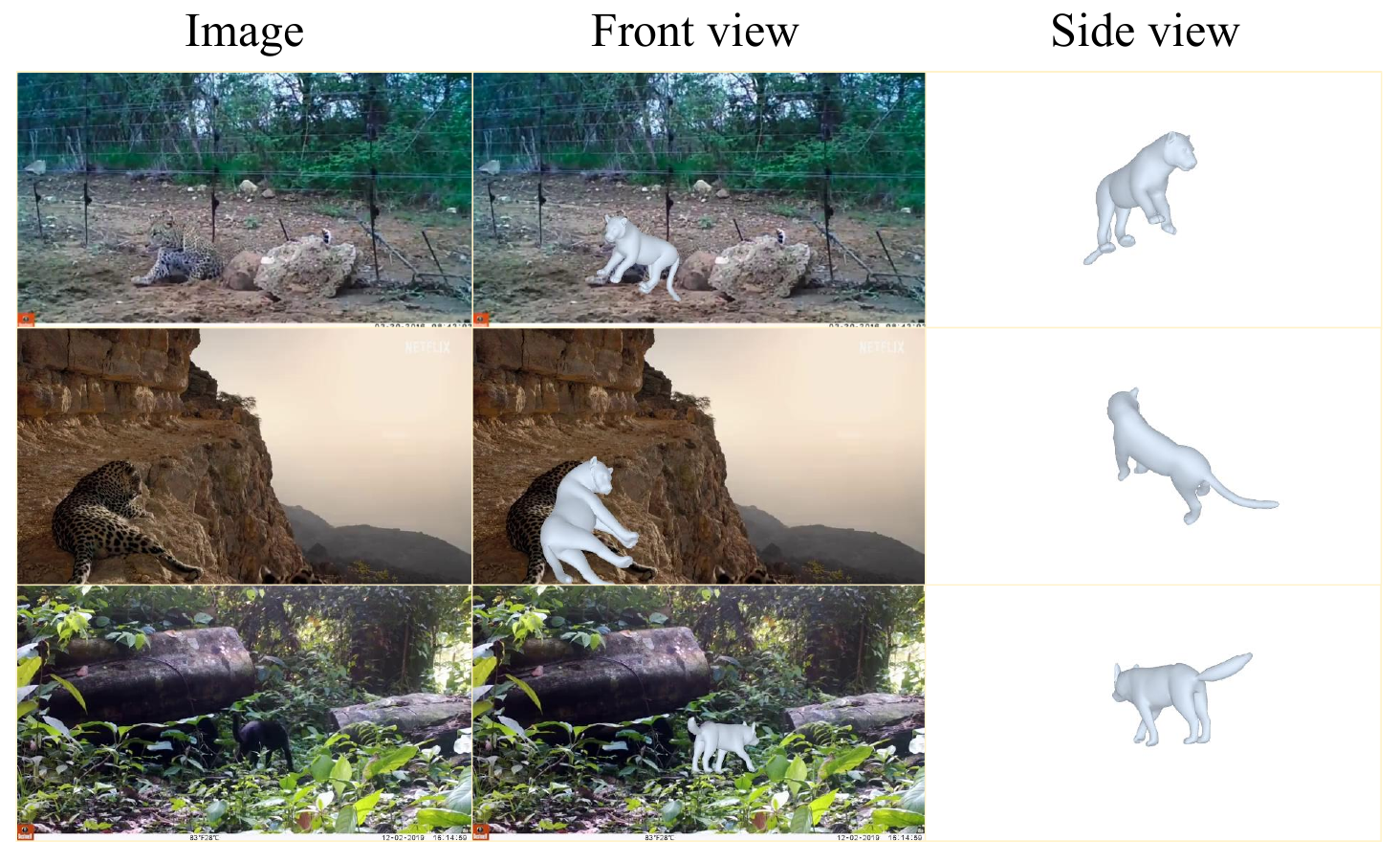}
    \caption{\textbf{Failure cases.} }
    \label{fig:failure_cases}
\end{figure*}

We provide failure cases in Fig.~\ref{fig:failure_cases}. Although AniMer demonstrates strong robustness, it can fail in certain scenarios. For example, large-scale occlusion (first row), extreme poses (second row) and excessively blurred images (third row) can lead to large reconstruction errors.

Our framework is based on SMAL, which is suitable for most quadrupedal animals. However, animals such as mice, fish, and birds cannot be represented using SMAL. As a result, we plan to adapt AniMer to accommodate a broader range of animal species in the future.

In addition, with the advent of large-scale synthetic datasets (e.g., GenZoo~\citep{niewiadomski2024generative}), we will further explore the performance of dataset scaling and contrastive learning on these extensive datasets.

\end{document}